# Use of Artificial Intelligence Techniques / Applications in Cyber Defense

Ensar Şeker
NATO CCD COE
Tallinn, Estonia
ensar.seker@ccdcoe.org

*Abstract*–Nowadays, considering the speed of the processes and the amount of data used in cyber defense, it cannot be expected to have an effective defense by using only human power without the help of automation systems. However, for the effective defense against dynamically evolving attacks on networks, it is difficult to develop software with conventional fixed algorithms. This can be achieved by using artificial intelligence methods that provide flexibility and learning capability. The likelihood of developing cyber defense capabilities through increased intelligence of defense systems is quite high. Given the problems associated with cyber defense in real life, it is clear that many cyber defense problems can be successfully solved only when artificial intelligence methods are used. In this article, the current artificial intelligence practices and techniques are reviewed and the use and importance of artificial intelligence in cyber defense systems is mentioned. The aim of this article is to be able to explain the use of these methods in the field of cyber defense with current examples by considering and analyzing the artificial intelligence technologies and methodologies that are currently being developed and integrating them with the role and adaptation of the technology and methodology in the defense of cyberspace.

## I. INTRODUCTION

Nowadays, the cyber area has reached the zettabyte age together with all the data. The emergence of new technologies and services, as well as trillions of devices and zettabytes, have left the cyber security experts with new threats and weaknesses, as well as addressing old threats and security vulnerabilities. In addition, the complex threats with technological developments and advances in cyber, created new cyber defense requirements for more rational solutions to occur.

The advent of new technologies in cyber defense systems has become critical at the expense of the technological escalation of threats to these systems. Increasing emphasis on artificial intelligence in cybernetic defense amongst new technologies contributes to the earlier response of cyber-attacks.

The basic problem to be addressed in the use of artificial intelligence in the cyber defense area is how artificial intelligence methodologies should be developed and adhered to in order to reduce the human factor, which is considered to be the weakest link of the cyber defense, while existing technologies are not in the desired level.

In this study, first of all, artificial intelligence is discussed, what is meant by this concept, the distinctions, the difficulties in developing this technology are mentioned. In later chapters, it was attempted to explain how the developed artificial intelligence techniques and methods that are used are used in cyber defense systems, which constitute of the main topic of this article. The use of artificial intelligence is especially important in early warning, intrusion detection and prevention systems. Again, examine the existing artificial intelligence applications that are used for cyber defense, will contribute in increasingly greater prominence to scientific studies.

## II. ARTIFICIAL INTELLIGENCE (AI)

Artificial intelligence is a computer science that deals with the development of systems and software that are intelligent, capable of performing tasks that are normally done by humans - equally or sometimes better. Many developed cognitive artificial intelligence methods (pattern recognition, cognitive intelligence, neural networks, intelligent agents, artificial immune systems, machine learning, data mining, fuzzy logic, heuristics, etc.) are actively used in the cyber field.

Artificial intelligence based on how to think like how a human thinks besides solitary thinking is called weak artificial intelligence [1], although artificial intelligence, which is called strong artificial intelligence, may exhibit movements like a human being. This distinction can also be made in some sources (thinking / acting, human / logical) [13]. Artificial Intelligence with real simulations (strong artificial intelligence) or systems that rely on the question of how a AI can thinks like a human being are not yet available. It is known that it is a very difficult problem to solve such an artificial intelligence that perceives events as a human being and reacts to them. The best example of weak artificial intelligence is Deep Blue, developed for chess games. Despite, being a good chess player, Deep Blue cannot pretend to be a human being. Apart from these two categories, there are examples of artificial intelligence developed between weak and strong as a third category. While these systems can be used as a guide for reasoning as a human being, they cannot be guided by an excellent modeling goal. One of the best examples of such a system is IBM Watson, which was developed by IBM. Watson is a question-and-answer computer system that can responds natural questions. It has the ability to recognize patterns in the text and combine matching patterns based on evidence [2, 3]. Another example related to the subject is the Google Brain developed by Google. Google Brain is an artificial intelligence research project based on deep learning. Google Brain's work structure has a brain-like work structure because it is inspired by the real structure of the brain. Deep learning systems based on the behavior of neurons function with learning layers for tasks such as painting and speech recognition [4].

Another distinction for artificial intelligence is general artificial intelligence and narrow artificial intelligence. General AI is designed for a general purpose is called and has no specific task, whereas artificial intelligence designed for a special purpose uses narrow artificial intelligence [13].

III. ARTIFICIAL INTELLIGENCE IN CYBER DEFENCE

Cyber defense is defending network of the critical infrastructure, the information security of the institutions and organizations, the government and the State departments, together with all other networks evaluated within the scope of national security, to eliminate possible threats of cyber-attacks and threats, and to reduce damage and losses caused by them, as a part of defense mechanisms. The cyber defense focuses on preventing infestation and / or threats from being detected and prevented in time, thus not altering infrastructure and / or data. With the complexity of the cyber-attacks as well as their advanced technologies, cyber defense is essential for most institutions and organizations to ensure the protection of sensitive information, data and assets. It also includes issues such as sensitive data about aggressors and degradation of the environment in which the assets are located, critical locations and assessing and understanding information, increasing the capacity of attack detection and reaction and response, and recognizing the ways, methods and areas that attackers can attack by technical analysis. In addition to the cyber security contributions, the cyber defense advocacy has also contributed to the development of security strategies, the most effective use of resources and, in particular, the effectiveness of security-related resources and expenditure at critical points.

As mentioned, avoiding cyber-attacks and avoiding cyber threats are the cornerstone of cyber defense. However, it is not possible to completely eliminate these attacks and threats. It is crucial to give the fastest response to these attacks and threats and to reduce the most possible damages. Existing security software databases and algorithms have a limited capacity and capability and often fail to cope with the rapid evolution and change of new threat vectors. Artificial intelligence algorithms designed in an intelligent security system have the potential to identify and respond to threats, even changing threats.

Nowadays, information security solutions are usually divided into two categories: analyst oriented or uncontrolled machine learning focused. Using uncontrolled machine learning to detect rare or abnormal patterns can increase the detection of new attacks. However, more false positives and warnings may also trigger. This requires a significant amount of analysis effort first to investigate the accuracy of these false positives. Such faulty alarms can cause alarm fatigue and insecurity, and over time, they may return to analytically focused solutions and lead to weaknesses associated with them why could it be. The IT security industry faces three major challenges, each of which can be addressed by machine learning solutions [5]:

- Lack of tagged data: Many organizations lack the ability to use tagged examples of pre-existing attacks and controlled learning models.
- Constantly evolving attacks: While supervised learning models are possible, attackers can override models by constantly changing their behavior.
- Limited Time and Budget for Research or Investigation Applying to investigate attacks on analysts is a costly and time consuming process.

Solutions that can overcome these difficulties should help the early stages of new and evolving attacks that will help analysts to use their time effectively, and to reduce the reaction times between attack detection and attack prevention, with extremely low false positive rates. A cyber security platform, developed by MIT based on Artificial Intelligence, called AI2, can provide these solutions [5]. Figure 1 shows about the AI2.

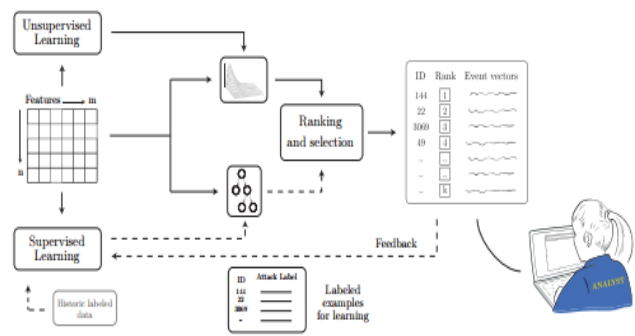

Figure 1. AI2 [5]

*A. Early Warning, Intrusion Detection & Prevention Systems in Cyber Defence*

Possible cyber defense system should provide at least three levels of cyber security. First level, identity and authentication, cryptographic protection, access control, auditing, network filtering, etc. as well as traditional static cyber defense mechanisms. The second level includes proactive cyber defense mechanisms such as information gathering, security assessment, network status monitoring, and attack. The third level refers to the holistic assessment of the network situation, the management of the cyber defense that fulfills the adaptation of the appropriate or optimal defense mechanisms and their adaptation [6].

Early warning, intrusion detection and prevention systems, which also contain artificial intelligence technologies, play an important role in the provision of such cyber security levels.

Early warning systems (EWS) are used to protect against cyber attacks and to respond as soon as possible. However, due to the new cyber threat level developed with new technologies, EWS based on traditional and pure packet inspection is also imroved. New EWS architecture that can collects, analyzes and correlates data. At the same time, it almost instantaneously perceives, analyzes and responds to threat models in real time . This need is evidenced by the development of virtual sensors, the sophisticated correlation of data, new logic models for network behavior analysis, learning algorithms and the development of new approaches and concepts that can provide scalability, reliability and resilience, especially in IPv6 networks [7]. Figure 2 shows typical intrusion detection and prevention system.

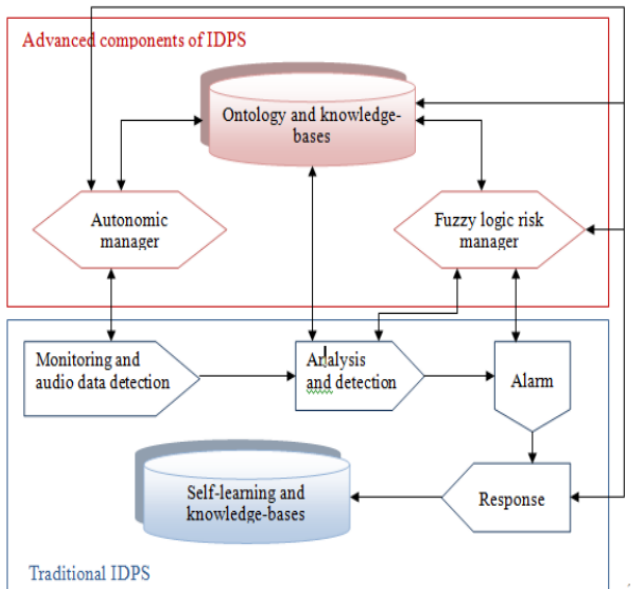

Figure 2 – Typical Intrusion Detection and Prevention Systems [8]

The purpose of using AI in early warning and intrusion detection is to develop an intelligent assist system to detect attacks from the Internet as early as possible on both local area networks and wide area networks. Within this framework, commonly used Internet protocols such as FTP, SMTP and HTTP should be considered, as well as newer protocols such as SOAP. One of the projects developed in this regard is the FIDeS project. The FIDeS project focuses on more assistance (such as concrete instructions in case of an attack) instead of just intrusion detection. For this purpose, various Artificial Intelligence-based methods are used, such as declarative information representation, statement generation and cognitive aids. It is designed to support the system security expert in analyzing attacks and developing counter attacks [9]. FIDeS system architecture is shown on Figure 3.

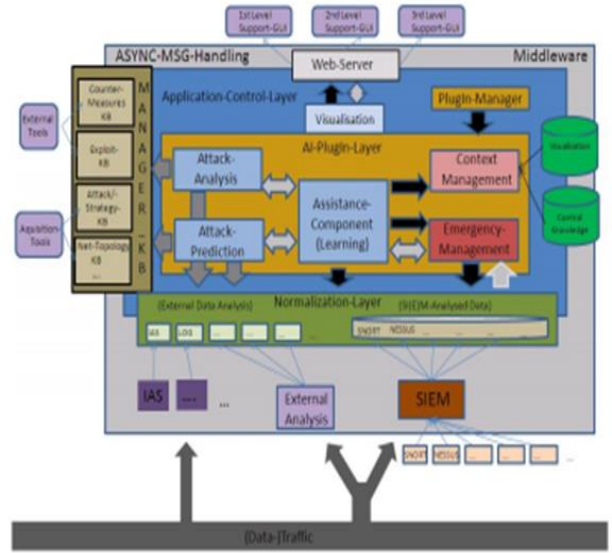

Figure 3 – FIDeS System Architecture [9]

It is expected that the intrusion detection and prevention system will have certain features for effective cyber defense against serious cyber-attacks. Some of these features are [10];
- To be able to perform real-time intrusion detection while the cyber-attack is ongoing or immediately afterwards,
- To minimize false positive alarms,
- To minimize human surveillance and ensure continuity of operations,
- To provide recoverability of the system against loss which may occur in the system due to accident or attack,
- Having the ability to self-control for the detection of attempts by attackers to make changes in the system,
- To comply with the security policies of the monitored system and
- To ensure compatibility with system changes over time and user behavior.

*B. Artificial Intelligence Applications for Cyber Defence*

Traditional hard-wired logic is ineffective to combat dynamically evolving cyber attacks. Therefore, more innovative approaches are needed, such as the use of artificial methods and practices that provide flexibility and learning skills, especially in cyber defense [11, 12].

The general problem of being able to simulate the intelligence has been simplified by identifying the sub-problems that have certain characteristics or abilities that an intelligent system should exhibit. Some of these sub-problems are; [13, 14]:
  I. Inference, logic, problem solving (embedded agents, neural networks, statistical approaches to artificial intelligence );
  II. Information presentation (ontology);

III. Planning (multiple agent planning and collaboration);
IV. Learning (machine learning);
V. Natural Language Processing (information acquisition - text review, machine translation);
VI. Movement and Manipulation (navigation, localization, mapping, motion planning);
VII. Perception (speech recognition, face recognition, object recognition);
VIII. Social Intelligence (empathy simulation);
IX. Creativity (artificial intuition, artificial imagination); and
X. General Intelligence (Strong Artificial Intelligence).

Considering the cyber defense, existing Artificial Intelligence methods and architectures can be divided into the following categories;

*1) Neural Nets:* Neural nets have a long history that began with the discovery of "perceptron" by Frank Rosenblatt in 1957. In machine learning, perceptron is an algorithm developed for supervisory learning of binary classifiers (functions that determine whether the input represented by vector numbers belongs to a particular class). One of the most popular elements of these neural networks is artificial neurons [15, 16]. A small number of perceptron that work together can learn to solve problems. But neural nets can consist of a large number of artificial neurons. Neural nets consisting of a large number of artificial neurons can provide massive parallel learning and decision-making functionality. The most prominent feature of these networks is their operational speed. They are well suited for pattern recognition, learning, classification, and attack response. Both hardware and software can be applied [17].

Neural nets are also suitable for intrusion detection and prevention [18, 19, 20, 21]. Scientific studies and suggestions have been made to use these nets in DoS detection [22], computer worm detection [23], spam detection [24], zombie detection [25], malware classification [26] and forensic investigations [27].

One reason why neural nets are popular in cyber defense is the high speeds that can be implemented in hardware and used in graphics processors. Third generation neural net - applications of spiking neural nets that mimic biological neurons - more realistically are among the new developments in neural net technology. Systems provided by Field Programmable Gate Arrays (FPGAs), which allow rapid development of neural nets and adapt to changing threats, provide significant contributions to cyber defense [12].

*2) Expert Systems:* Artificial Intelligence expert systems are the most used tools. An expert system is software that is used to find answers to questions posed by a user or other software in the activity areas in some applications. They can be used directly to support decisions in areas such as finance, medical diagnostics or cyberspace. There are a variety of expert systems ranging from small technical diagnostic systems to complex, very large and sophisticated hybrid systems to solve problems. Conceptually, an expert system includes a knowledge base in which expert knowledge about a particular application domain is stored. In addition to this information substructure, it also has an inference engine and additional information about the situation to get answers based on this knowledge. The empty information base and the extraction engine are collectively called the expert system shell - must be filled with information so that it can be used. The expert system shell can be supported by the software to add information to the knowledge base and must be extensible for user interactions and other programs that can be used in hybrid expert systems. Developing an expert system means firstly choosing an expert system shell and adaptation. Secondly filling in expert knowledge and knowledge base with information. The second step is much more complicated than the first step and takes much more time than originally.

An example, for expert systems that can be used in cyber defense, is safety planning [28]. An expert system using this area greatly simplifies the task of selecting security measures and provides guidance on how best to use limited resources. In addition, the use of expert systems in attack detection is used longstanding [29, 30].

*3) Intelligent Agents:* Intelligent agents are software components that have some characteristics with intelligent behaviors makes (proactivity, agent communication, language understanding and response) them special. These software components have planning, variability and deep thought capabilities. Software agents have been adopted as a concept in software engineering where they are thought of as objects that use proactive and agent communication language. However, when agents and objects are compared, they can be shown as differences in which objects can be passive and do not require any language to speak (although they accept the well-defined syntax) [12].

There are studies simulating how intelligent agents are effective against DDoS attacks in the use of cyber defense [31, 32]. In some of these studies stated that it is possible to develop a "cyber police" consisting of mobile intelligent agents after solving some legal and commercial problems [33]. There are also hybrid multi-agent and neural network-based intrusion detection systems [34] and agent-based distributed intrusion detection systems [35].

*4) Search:* Search is found in almost every intelligent program in a variety of shapes and formats, and its efficiency is usually critical for the performance of the entire program. If additional information is available to guide the search, while fulfilling the requirements for a solution, the search activity can be significantly improved. Artificial Intelligence has developed many search methods. Although it is used in many software, it is generally not seen as the use of Artificial Intelligence. For example, in dynamic programming [36, 37] that is used to solve optimal security problems, search software is embedded and does not appear to be an Artificial Intelligence application. Andor trees, αβ-search, minimax search and stochastic search are widely used in game software and are useful for decision making for cyber defense. The αβ-search algorithm originally developed for computer chess games that is very successful in problem solving and especially in evaluating and determining the best possible actions of the two attacks. This algorithm, which uses estimates of least winning and losing most, accelerates the search by ignoring a large number of choices.

*5) Learning:* Learning improves an information system by expanding or reorganizing its knowledge base or by improving the inference engine [38]. Machine learning involves new methods of calculating information to obtain new ways of organizing new knowledge, new skills and existing knowledge. Learning problems vary greatly from simple parametric learning (learning the values of certain parameters and the complex forms of symbolic learning such as learning of concepts, language constructs, functions, and even behavior learning).

Artificial Intelligence offers both supervised learning (teacher learning) and methods for uncoordinated learning. Uncontrolled learning is particularly useful when large amounts of data are available, and this method is widespread in cyber defense where large diaries can be collected. Data mining was originally from Artificial Intelligence untrained learning [18, 39].

An elite class of learning has been created by parallel learning algorithms that are suitable for execution on parallel hardware. These learning methods are represented by genetic algorithms and neural networks. Genetic algorithms and fuzzy logic methods have been used in cyber defense, for example, in threat detection systems [40].

*6) Constraint Solving:* Constraint solving or constraint satisfaction is a technique developed using Artificial Intelligence [41] to solve problems presented (logical expressions, tables, equations, inequalities) by giving a set of constraints on the solution. The solution to a problem is a collection of values that satisfy all constraints (a set). In fact, there are many different constraint-resolution techniques depending on the nature of the constraints (eg constraints on the final sets, functional constraints, rational trees). At a very abstract level, almost any problem can be presented as a constraint saturation problem. The solution of these problems is generally very difficult because of the need for many searches. Constraint solving can be used in logic analysis as well as in state analysis and decision support [42, 43].

## IV. CONCLUSION AND FUTURE WORK

The US Department of Defense has issued five pillars for the cyber warfare environment in its published cyber strategy document. These are the pillars [44];

i. The cyber area carries elements similar to other battlefields as a new field of activity and should therefore be adopted as a new battlefield after air, land, sea and space.
ii. Passive proactive defenses should be substituted for defense.
iii. Critical Infrastructure Protection (CIP) concept should be adhered to ensure the protection of infrastructure.
iv. Early threat detection systems for providing the ability to identify should be part of defensive structures for joint defence.
v. Advantages of technological change have to be preserved, developed and new technologies (especially artificial intelligence) have been adapted to the cyber defense systems.

The fifth pillar from these five basic principles is particularly important. In cyber space, the data values has reached that can not be categorized and managed by people, the speed of technological developments and more complicated and sophisticated cyber threats that come along with these developments. It is critically important and necessary to detect and prevent as early as possible to reduce the damages of these threats may cause. Artificial intelligence became an indispensable element of cyber defence. That's why understanding and adapting AI solutions for cyber defence draws serious attention, especially for researchers.

Future work will focus on AI applications and solutions for critical infrustructure. Another future work will be on the integration and use of artificial echoes that are desired to be used for defending against cyber attacks on cyber defense exercises.

## V. REFERENCES

[1] S. Dilek, H. Çakır, M. Aydın, "Applications of Artificial Intelligence Techniques to Combating Cyber Crimes: A Review", 2015.

[2] R. High, "The Era of Cognitive Systems: An Inside Look at IBM Watson and How it Works", IBM, 2012.

[3] J. E. Kelly, "Computing, Cognition and the Future of Knowing", IBM, 2015.

[4] J. Dean, "Large-Scale Deep Learning for Intelligent Computer Systems", 2016.

[5] K. Veeramachaneni, I. Arnaldo, A. Cuesta-Infante, V. Korrapati, C. Bassias, K. Li, "AI2: Training a Big Data Machine to Defend", IEEE International Conference on Big Data Security in New York City, 2016.

[6] I. Kotenko, "Multi-agent Modelling and Simulation of Cyber-Attacks and Cyber-Defense for Homeland Security", IEEE International Workshop on Intelligent Data Acquisition and Advanced Computing Systems: Technology and Applications, 2007.

[7] M. Golling, B. Stelte, "Requirements for a Future EWS – Cyber Defence in the Internet of the Future", 3rd International Conference on Cyber Conflict, CCD COE, 2011.

[8] A. Patel, M. Taghavi, K. Bakhtiyari, J. Celestino Junior, "An Intrusion Detection and Prevention System in Cloud Computing: A Systematic Review", Journal of Network and Computer Applications, Elsevier, 2013.

[9] S. Edelkamp, C. Elfers, M. Horstmann, M. S. Schröder, K. Sohr, T. Wagner, "Early Warning and Intrusion Detection based on Combined AI Methods", 2009.

[10] A. Patel, Q Qassim, Z. Shukor, J. Nogueira, J Júnior, C. Wills, "Autonomic Agent-Based Self-Managed Intrusion Detection and Prevention System, Proceedings of the South

African Information Security Multi-Conference", Port Elizabeth, South Africa, 2010.

[11] J. Helano, M. Nogueira, "Mobile Intelligent Agents to Fight Cyber Intrusions", the International Journal of Forensic Computer Science, 2006.

[12] E. Tyugu, "Artificial Intelligence in Cyber Defense", 3rd International Conference on Cyber Conflict, 2011.

[13] J. S. Russell, P. Norvig, "Artificial Intelligence: A Modern Approach", 2nd edition, Upper Saddle River, Prentice Hall, New Jersey, USA, 2003.

[14] G. Luger, W. Stubblefield, "Artificial Intelligence: Structures and Strategies for Complex Problem Solving", Addison Wesley. 2004.

[15] F. Rosenblatt. "The Perceptron - A Perceiving and Recognizing Automaton", Cornell Aeronautical Laboratory, 1957.

[16] Y. A. Freund, R. E. Schapire, "Large Margin Classification Using the Perceptron Algorithm, Machine Learning", 37(3):277-296, 1999.

[17] G. Klein, A. Ojamaa, P. Grigorenko, M. Jahnke, E. Tyugu, "Enhancing Response Selection in Impact Estimation Approaches", Military Communications and Information Systems Conference (MCC), Wroclaw, Poland, 2010.

[18] J. Bai, Y. Wu, G. Wang, S. X. Yang, W. Qiu, "A Novel Intrusion Detection Model Based on Multilayer Self-organizing Maps and Principal Component Analysis, Advances in Neural Networks", ISNN Springer Berlin Heidelberg, 2006.

[19] F. Barika, K. Hadjar, N. El-Kadhi, "Artificial Neural Network for Mobile IDS Solution", Security and Management, 2009.

[20] D. A. Bitter, T. Elizondo, "Application of Artificial Neural Networks and Related Techniques to Intrusion Detection", IEEE World Congress on Computational Intelligence, CCIB, Barcelona, Spain, 2010.

[21] R. I. Chang, L. B. Lai, W. D. Su, J. C. Wang, J. S. Kouh, "Intrusion Detection by Backpropagation Neural Networks with Sample-query and Attribute-query", International Journal of Computational Intelligence Research, 2007.

[22] B. Iftikhar, A. S. Alghamdi, "Application of Artificial Neural Network in Detection of DOS Attacks", Proceedings of the 2nd international Conference on Security of Information and Networks. New York, NY, 2009.

[23] D. Stopel, Z. Boger, R. Moskovitch, Y. Shahar, and Y. Elovici, "Application of Artificial Neural Networks Techniques to Computer Worm Detection", International Joint Conference on Neural Networks, 2006.

[24] C. H. Wu, "Behavior-based Spam Detection Using a Hybrid Method of Rule-based Techniques and Neural Networks", Expert Systems with Applications, 2009.

[25] P. Salvador, et al., "Framework for Zombie Detection Using Neural Networks", Fourth International Conference on Internet Monitoring and Protection, 2009.

[26] M. Shankarapani, K. Kancherla, S. Ramammoorthy, R. Movva, S. Mukkamala, "Kernel Machines for Malware Classification and Similarity Analysis", IEEE World Congress on Computational Intelligence. Barcelona, Spain, 2010.

[27] B. Fei, J. Eloff, M. S. Olivier, H. Venter, "The Use of Self-organizing Maps of Anomalous Behavior Detection in a Digital Investigation", Forensic Science International, 2006.

[28] J. Kivimaa, A. Ojamaa, E. Tyugu, "Graded Security Expert System", Springer, 2009.

[29] D. Anderson, T. Frivold, A. Valdes, "Next-generation Intrusion Detection Expert System (NIDES)", SRI International, Computer Science Lab, 1995.

[30] T. F. Lunt, R. Jagannathan, "A Prototype Real-Time Intrusion-Detection Expert System", IEEE Symposium on Security and Privacy, 1988.

[31] I. Kotenko, A. Ulanov, "Multi-Agent Framework for Simulation of Adaptive Cooperative Defense Against Internet Attacks", International Workshop on Autonomous Intelligent Systems: Agents and Data Mining, Springer.

[32] I. Kotenko, A. Konovalov, A. Shorov, "Agent-Based Modeling and Simulation of Botnets and Botnet Defence", Conference on Cyber Conflict, CCD COE Publications, Tallinn, Estonia, 2010.

[33] B. Stahl, D. Elizondo, M. Carroll-Mayer, Y. Zheng, K. Wakunuma, "Ethical and Legal Issues of the Use of Computational Intelligence Techniques in Computer Security and Computer Forensics", IEEE World Congress on Computational Intelligence, Barcelona, Spain, 2010.

[34] E. Herrero, M. Corchado, A. Pellicer, A. Abraham, "Hybrid Multi Agent-neural Network Intrusion Detection with Mobile Visualization", Innovations in Hybrid Intelligent Systems, 2007.

[35] V. Chatzigiannakis, G. Androulidakis, B. Maglaris, "A Distributed Intrusion Detection Prototype Using Security Agents". HP OpenView University Association, 2004.

[36] J. Kivimaa, A. Ojamaa, E. Tyugu, "Pareto-Optimal Situation Analysis for Selection of Security Measures", MilCom, 2008.

[37] J. Kivimaa, A. Ojamaa, E. Tyugu, "Managing Evolving Security Situations", MilCom, 2009.

[38] P. Norvig, S. Russell, "Artificial Intelligence: Modern Approach", Prentice Hall, 2000.

[39] V. K. Pachghare, P. Kulkarni, D. M. Nikam, "Intrusion Detection System using Self Organizing Maps", International Conference on Intelligent Agent & Multimedia Agent Systems, 2009.

[40] R. Hosseini, J. Dehmeshki, S. Barman, M. Mazinani, S. Qanadli, "A Genetic Type-2 Fuzzy Logic System for Pattern Recognition in Computer Aided Detection Systems", IEEE World Congress on Computational Intelligence. Barcelona, Spain, 2010.

[41] B. Mayoh, E. Tyugu, J. Penjam, "Constraint Programming", NATO ASI Series, Springer Verlag. 1994.

[42] I. Bratko, "PROLOG Programming for Artificial Intelligence", Addison-Wesley, 2001.

[43] X. Ou, "A Logic-programming Approach to Network Security Analysis", PhD Thesis, Princeton University, 2005.

[44] "US Department of Defense Cyber Strategy", US DoD, April 2015.